\RequirePackage{fix-cm}
\documentclass[twocolumn]{svjour3}          %
\smartqed  %
\usepackage{graphicx}
\usepackage{mathptmx}      %
\usepackage{latexsym}

\usepackage{algorithm}
\usepackage{algpseudocode}
\usepackage{amsfonts}       %
\usepackage{amsmath}
\usepackage{amssymb}
\usepackage{array}
\usepackage{arydshln}
\usepackage{bbding}
\usepackage{bm}
\usepackage{booktabs}       %
\usepackage{colortbl}
\usepackage{dsfont}
\usepackage{etoc}           %
\usepackage[pagebackref,breaklinks,colorlinks,citecolor=blue]{hyperref}
\usepackage{makecell}
\usepackage{microtype}      %
\usepackage{multirow}       %
\usepackage{natbib}         %
\usepackage{nicefrac}       %
\usepackage{orcidlink}      %
\usepackage{pifont}
\usepackage{tcolorbox}
\usepackage{url}
\usepackage{wrapfig}
\usepackage{caption}

\definecolor{DarkOrange}{RGB}{225,120,20}
\definecolor{Amethyst}{RGB}{153,102,204}

\newcommand{\citeps}[1]{{{\citep{#1}}}}
\newcommand{\cites}[1]{{{\citep{#1}}}}
\newcommand{\emails}[1]{E-mail:~{\small #1}}

\newcommand{\vvb}[1]{\bm{#1}}

\newcommand{\eg}{\textit{e.g.}}

\newcommand{\cmark}{\textcolor{green!60!black}{\large\ding{51}}} %
\newcommand{\xmark}{\textcolor{red!70!black}{\large\ding{55}}}   %

\definecolor{DarkOrange}{RGB}{180,0,180}
\newcommand{\rrev}[1]{{#1}} %

\journalname{International Journal of Computer Vision Under Reviewing}
\begin{document}

\title{SEE: See Everything Every Time - Adaptive Brightness Adjustment for Broad Light Range Images via Events}
\subtitle{A Large Scale Dataset and a Novel Lightweight Framework}

\titlerunning{SEE: See Everything Every Time}        %
\authorrunning{Yunfan Lu \textit{et. al.}}

\author{ \small
Yunfan~Lu$^1$ \and Xiaogang~Xu$^2$ \and Hao~Lu$^1$ \and Yanlin~Qian$^3$ \and Pengteng~Li$^1$ \and Huizai~Yao$^1$ \and Bin~Yang$^4$ \and Junyi~Li$^1$ \and Qianyi~Cai$^1$ \and Weiyu~Guo$^1$ \and Hui~Xiong~*$^1$}

\institute{
Yunfan Lu \at
$^1$~AI Thrust, HKUST(GZ), China \\
\emails{ylu066@connect.hkust-gz.edu.cn}
\and
Xiaogang Xu \at
$^2$~CUHK, HongKong, China \\
\emails{xiaogangxu00@gmail.com}
\and
Yanlin Qian \at
$^3$~DJI~/~Hasselblad, \\
\emails{honza.qian@dji.com}
\and
Hao Lu,  Pengteng Li, Huizai Yao, \\Junyi Li, Qianyi Cai and Weiyu Guo \at
$^1$~AI Thrust, HKUST(GZ), China \\
\and
Bin Yang \at
$^4$~Aalborg University, Denmark \\
\emails{byang@cs.aau.dk}
\and
\textbf{Hui Xiong *} \at
$^1$~AI Thrust, HKUST(GZ), China \\
\textit{Corresponding author} \\
\emails{xionghui@ust.hk}
}

\date{Received: date / Accepted: date}

\maketitle

\begin{abstract}
Event cameras, with a high dynamic range exceeding $120dB$, significantly outperform traditional embedded cameras, robustly recording detailed changing information under various lighting conditions, including both low- and high-light situations.
However, recent research on utilizing event data has primarily focused on low-light image enhancement, neglecting image enhancement and brightness adjustment across a broader range of lighting conditions, such as normal or high illumination.
Based on this, we propose a novel research question: \textbf{how to employ events to enhance and adaptively adjust the brightness of images captured under broad lighting conditions?}
To investigate this question, we first collected a new dataset, SEE-600K, consisting of 610,126 images and corresponding events across 202 scenarios, each featuring an average of four lighting conditions with over a 1000-fold variation in illumination.
Subsequently, we propose a framework that effectively utilizes events to smoothly adjust image brightness through the use of prompts.
Our framework captures color through sensor patterns, uses cross-attention to model events as a brightness dictionary, and adjusts the image's dynamic range to form a broad light-range representation (BLR), which is then decoded at the pixel level based on the brightness prompt.
Experimental results demonstrate that our method not only performs well on the low-light enhancement dataset but also shows robust performance on broader light-range image enhancement using the SEE-600K dataset.
Additionally, our approach enables pixel-level brightness adjustment, providing flexibility for post-processing and inspiring more imaging applications.
The dataset and source code are publicly available at:~{\url{https://github.com/yunfanLu/SEE}}.
\end{abstract}

\begin{figure*}[t!]
\includegraphics[width=0.99\linewidth]{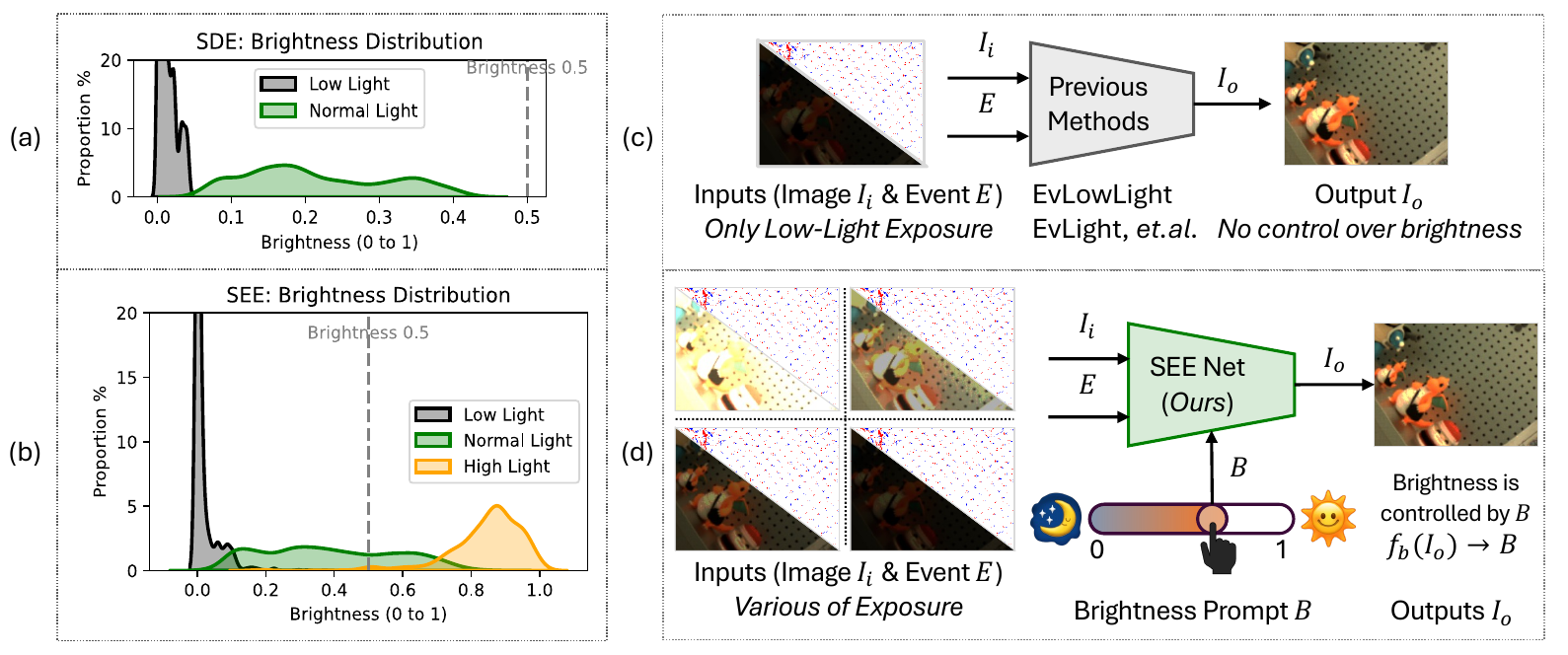}
\caption{{(a) and (b): Brightness distributions of the SDE dataset (\textbf{0$\sim$0.45}, low to normal light) and our SEE-600K dataset (\textbf{0$\sim$1}, a broader light range). (c): Previous methods~\citep{liang2023coherent,liang2024towards} directly map low-light images to normal-light images. (d): Our SEENet accepts inputs across a broader brightness range and adjusts output brightness through prompts. $f_b$ refers to the function that calculates the brightness of an image.}}
\label{fig:0-cover-figure}
\end{figure*}

\section{Introduction}
Every day, from daylight to nighttime, the illuminance varies from about 100,000 \textit{lux} (bright sunlight) to approximately 0.1 \textit{lux} (starlight) ~\citeps{koshel2012illumination}.
Maintaining stable imaging under diverse natural lighting conditions is a significant challenge that energizes many embedded applications~\citep{lin2024embodied}.
To achieve this, a series of influential works have emerged, including automatic exposure~\citeps{bernacki2020automatic}, exposures correction~\citeps{yuan2012automatic}, low-light enhancement~\citeps{li2021low} and high dynamic range (HDR) imaging~\citeps{mccann2011art}. 
However, traditional cameras are limited by their imaging principle of synchronously capturing intensity values across the entire sensor~\citep{hasinoff2016burst,rebecq2019high}.
Consequently, these traditional methods find it difficult to capture imaging information under a wide range of lighting conditions at the input~\citeps{gehrig2024low,gallego2020event}.
If the exposure is inaccurate - over and under exposures - traditional embedded cameras lose the potential to restore images under complex lighting conditions due to limited bits-width and noise.
Unlike traditional cameras, event cameras ~\citep{gallego2020event} asynchronously record pixel-level changes in illumination, outputting the direction of intensity change (positive or negative) at each pixel with extremely high dynamic range (120 $dB$), which far exceeds the capability of traditional cameras in capturing various lighting intensity.

Research leveraging the events for image brightness enhancement can be divided into three categories.
\textbf{(1) event-based image reconstruction}, which aims to reconstruct images only from events.
\textit{However, these methods~\citeps{rebecq2019high,stoffregen2020reducing,wang2024revisit} rely solely on events, facing uncertainties during reconstruction, and the events usually contain noise, which leads to color distortion and limited capabilities of generalization.}
\textbf{(2) event-guided HDR imaging}
~\citeps{cui2024color,yang2023learning,messikommer2022multi}
, which targets to employ events to extend the dynamic range of images or videos to match human vision.
~\citet{cui2024color} introduced the first real-world dataset containing paired color events, low dynamic range images, and HDR images, with only includes 1,000 HDR image pairs.
~\citet{messikommer2022multi} used nine images with different exposures to synthesize an HDR image as the ground truth and utilized multi-exposure frames and events as inputs to generate an HDR image.
\textit{While HDR imaging aims to expand dynamic range, collecting HDR datasets is difficult, and these methods have not been evaluated for tasks like low-light enhancement or high-light restoration~\citep{tursun2015state,chen2024evlight++,liang2023coherent,liang2024towards,jayasuriya2023software}.}
\textbf{(3) event-guided low-light enhancement}
~\citeps{liang2023coherent,liu2023low,jiang2023event,liang2024towards}, 
which is designed to adjust low-light images to normal-light conditions through brightness adjustment and noise reduction.
\citet{liang2024towards} represents the latest research and proposed the first event-based low-light image enhancement dataset, SDE (see Fig.~\ref{fig:0-cover-figure}~(a)).
Prior to this, \citet{liang2023coherent,liu2023low,jiang2023event} explored using motion information from events and employed varying neural networks to improve the mapping from low-light images to normal-light ones, as shown in Fig.~\ref{fig:0-cover-figure}~(c).
\textit{However, these strategies only focus on the improvement of mapping ability for low-light inputs, limiting their capacity to adjust brightness across a broader range of lighting conditions, \eg, normal or high-light images.}
The fundamental reason lies in the fact that these methods rely on datasets (such as SDE) that control brightness using a neutral density filter, \textit{i.e.}, they create paired low-light and normal-light images with \textbf{only one} fixed lighting ratio between the two. 
This approach forces the model to learn a rigid multiplicative enhancement strategy that only works within the narrow bounds of low-light to normal-light transitions. 
Consequently, the methods trained on such datasets are unable to generalize well to broad lighting conditions, particularly those involving high-light scenarios or complex lighting variations. 
This limitation arises because the training data does not reflect the full diversity of lighting conditions encountered in real-world applications, where illumination can change dynamically and across a much wider range. Therefore, these methods lack the flexibility needed to adjust brightness across a broader range of lighting variations.

Furthermore, due to the uncertainty in the standard for normal-light image collection—as the normal-light images are relative to low-light images (as shown in Fig.~\ref{fig:0-cover-figure}~(a))—these methods introduce ambiguity during the training process because they can only map low-light images to normal-light ones based on a single set of low- and normal-light data pairs captured per scene.
Overall, current research focuses on low-light enhancement, neglecting image enhancement and processing under a wider range of lighting conditions.
Therefore, \textbf{\textit{how to use events to enhance and adjust the brightness of images across a broader range of lighting conditions}} becomes a more worthwhile research question.

To address this novel research question, we first formulate the imaging model for brightness adjustment (Sec.~\ref{sec:preliminary}) and define the new learning task. 
We aim to perceive lighting information from events, utilizing brightness prompts to convert this lighting information into images with a specific brightness.
In doing so, other image quality aspects (like sensor patterns, noise, and color bias) are taken into consideration.

To realize our proposed task, we first collected a new dataset by emulating each scene in different lighting conditions, covering a broader luminance range (Sec.~\ref{sec:see_dataset}), as shown in Fig.~\ref{fig:0-cover-figure} (b) and (d).
By capturing multiple lighting conditions per scene, we enable mappings across diverse illumination scenarios, providing rich data for model training.
To tackle the challenges of spatio-temporal alignment of video and event streams under various lighting conditions, we design a temporal alignment strategy relying on programmable robotic arms and inertial measurement unit (IMU) sensors.
As a result, we obtain a temporal registration error up to one millisecond and a spatial error at the sub-pixel level ($\sim 0.3$ \textit{pixel}).
Finally, we build a large-scale and well-aligned dataset containing 202 scenes, each with 4 different lighting conditions, summing up to 610,126 images and the corresponding event data. 
We term this dataset as SEE-600K, which supports learning the mappings among multiple lighting conditions.

Building on the SEE-600K dataset, we propose a compact and efficient framework, SEE-Net, for the proposed new task (Sec.~\ref{sec:method}). 
An event-aware cross-attention is used to enhance image brightness, and the brightness-related prompt is introduced for controlling the overall brightness.
This approach effectively captures and adjusts lighting across a broader range of illumination conditions, providing flexibility and precise control during inference. Despite the performance advantage, SEE-Net still remains effective, compact, and lightweight with only \textbf{1.9 \textit{M}} parameters.

\begin{figure*}[t!]
\centering
\includegraphics[width=\linewidth]{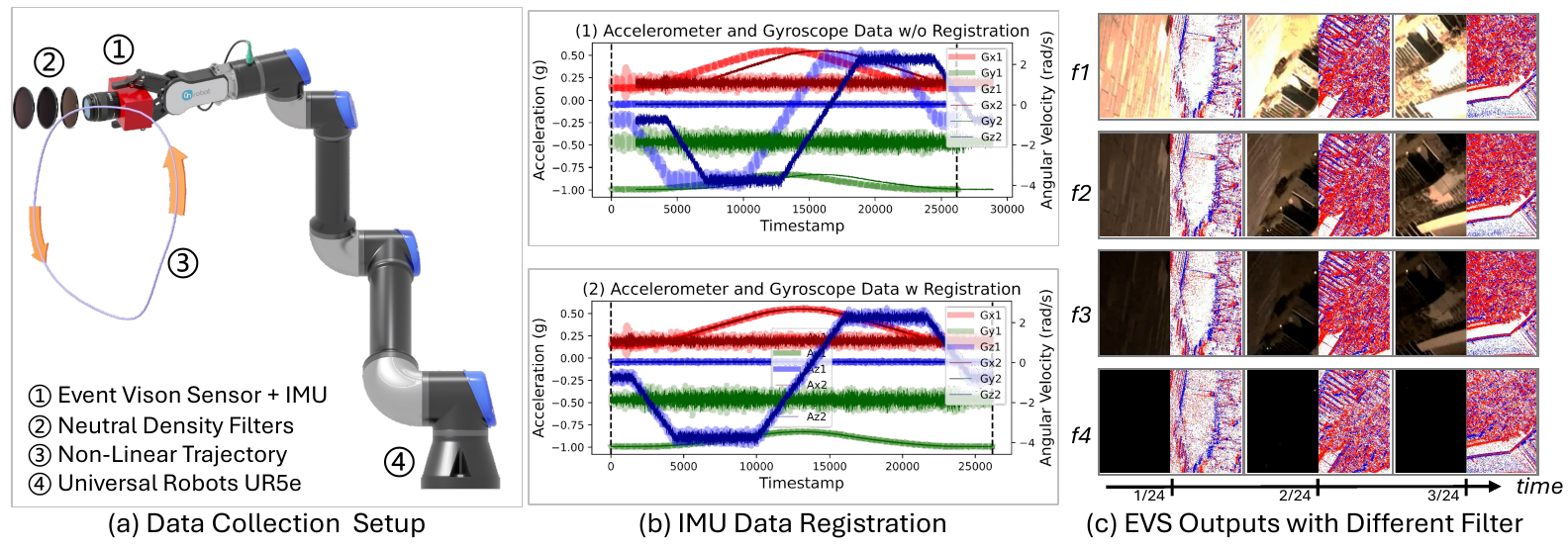}
\caption{{\textbf{(a) data collection setup:} Universal Robots UR5e arm replicates precise trajectories with an error margin of $0.03 mm$.
\textbf{(b) IMU data registration:} \textit{b (1)} shows unregistered IMU data, while \textit{b (2)} displays registered data after timestamp alignment.
\textbf{(c) EVS outputs with different filters:} \textit{f1} to \textit{f4} demonstrate the different ND filters, depicting various lighting levels.}
}
\label{fig:1-Arm-Release}
\end{figure*}

Our method has been evaluated on two real-world datasets, SDE~\citeps{liang2024towards} and SEE-600K. 
Quantitative results demonstrate that our framework fits well to a broader range of lighting conditions (Sec.~\ref{sec:experiments}). 
Furthermore, our framework allows for smooth brightness adjustment, providing precise exposure control.
Therefore, this flexibility significantly improves post-processing capabilities and enables potential applications in advanced imaging and processing tasks.

\section{Related Works\label{sec:related_works}}

\noindent\textbf{Frame-based:} These brightness enhancement methods aim to improve image quality under challenging illumination conditions.
Retinexformer~\citeps{cai2023retinexformer} and other Retinex-based frameworks~\citeps{zhang2021beyond,wu2022uretinex,fu2023you} decompose reflectance and illumination with complex training pipelines.
Other approaches, \eg, structure-aware models~\citeps{xu2023low,wang2023low}, utilize edge detection or semantic-aware guidance to achieve sharper and more realistic results.
Exposure correction strategies~\citeps{afifi2021learning,panetta2022deep,ma2020joint} target both overexposed and underexposed areas, leveraging multi-scale networks or perceptual image enhancement frameworks to synthesize correctly exposed images.
However, the reliance on RGB frames with limited bit depth, limits the adaptability to dynamic lighting conditions, making it difficult to handle a broader range of lighting scenarios.

\noindent\textbf{Event-based:} These methods focus on reconstructing images or videos exclusively from event data. For instance, \citet{duwek2021image} introduced a two-phase neural network combining convolutional neural networks and spiking neural networks, while \citet{pan2019bringing} proposed the event-based double integral model to generate videos. \citet{stoffregen2020reducing} enhanced event-based video reconstruction by introducing the new dataset. Additionally, \citet{liu2023sensing,wang2024revisit} developed a model-based deep network to improve reconstructed video quality.
However, these event-based approaches face challenges due to event data noise, often leading to color distortion and limited generalization.

\noindent\textbf{Event-guided:} These works are centered on enhancing images captured in low-light conditions. E.g., \citet{zhang2020learning} and \citet{liu2024seeing} recovered lost details in low-light environments by reconstructing grayscale images. Similarly, \citet{liang2023coherent} and \citet{liu2023low} improved low-light video enhancement by leveraging motion information from events to enhance multi-frame videos and integrate spatiotemporal coherence. Furthermore, \citet{jin2023event} and \citet{jiang2023event} utilized events to recover structural details and reconstruct clear images under near-dark situations. 
Most notably, \citet{liang2024towards} introduced the first large-scale event-guided low-light enhancement dataset, which is significant for the development of this field. 
While these methods use events for brightness changes and structural recovery in low-light conditions, they are limited to enhancing low-light images with single mapping and cannot handle brightness adjustments across a broader range of lighting conditions, including normal- and high-light.

\section{Preliminaries and New Task Definition\label{sec:preliminary}}
In this section, we formalize the physical model underlying our approach to enhance and adjust image brightness across a broader range of lighting conditions using events.
Imaging is fundamentally the process of capturing the light intensity of a scene, represented as a radiance field $L(t)$ varying over a preset slot $t$. 
The illumination intensities of light in daily life span a vast range, from 0.1~\textit{lux} (starlight) to $1e6$~\textit{lux} (direct sunlight). 
The goal of brightness adjustment is to recover or estimate $L(t)$ and tone-map it into an image that is visually suitable for human perception.

\begin{figure*}[t!]
\centering
\includegraphics[width=\linewidth]{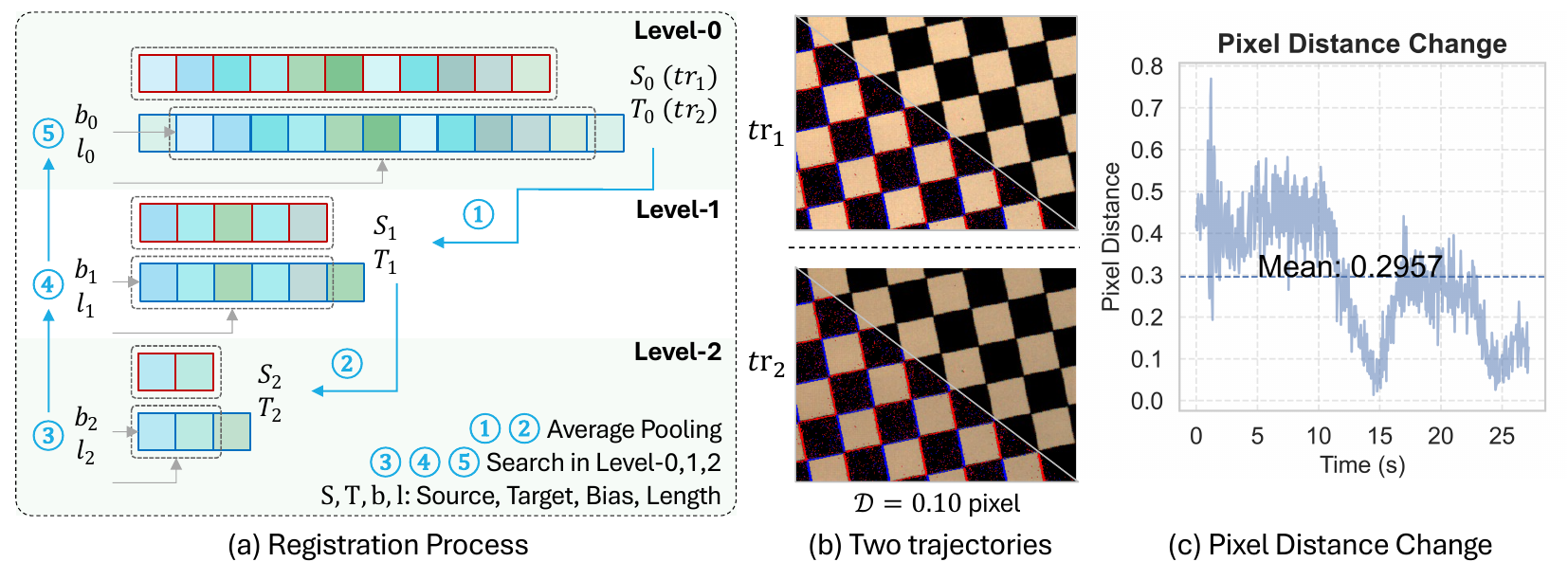}
\caption{{\textbf{(a) registration process:} Illustration of the multi-level registration process, showing how trajectories, $S$ and $T$, at various levels are iteratively aligned.
\textbf{(b) two trajectories:} The example of two aligned images captured along two trajectories.
\textbf{(c) pixel distance change:} Temporal distance of pixel between two registered videos, showing a mean alignment error of 0.2957 pixels over time.}
\label{fig:2-Reg-Vis-Release}}
\end{figure*}

Traditional cameras record light signals through exposure~\citeps{mendis1997cmos}.
This voltage is influenced by the Gaussian noise $N = \mathcal{N}(\mu, \sigma^2)$ ($\mu$ is the mean and $\sigma^2$ is the variance), and the photon shot noise $P = \mathcal{P}(k)$, where $k \propto L(t)$ is the number of photons, proportional to light intensity.
In low-light conditions, Gaussian noise becomes dominant, while in high-light conditions, photon shot noise becomes more significant.
These noise types influence the final value in the RAW image, 
simply represented as $I_{\text{raw}} \approx \mathcal{Q}( L(t) + P + N)$, where $\mathcal{Q}$ is the quantization function that converts the continuous voltage into discrete digital signals, typically ranging from 8 to 12 bits. 
The shape of the image $I_{\text{raw}}$ is $H\times W\times 1$, where $H$ and $W$ are the image resolution.
The RAW image is then further processed through image signal processing (ISP) $f_{\text{isp}}$, which includes multiple steps \eg, denoising, linear and non-linear transformations, resulting in a RGB image as $I_{\text{rgb}} = f_{\text{isp}}(I_{\text{raw}})$, with the shape of $H\times W\times 3$.
An accurate image exposure procedure recovers $I_{\text{rgb}}$ corresponding to $L(t)$, up to a high degree meeting the following three characteristics: 
\textbf{(1) accurate exposure}: The mean value of $I_{\text{rgb}}$ falls within the range $[0.4, 0.7]$~\citeps{mertens2009exposure}.
\textbf{(2) noise-free}: The influence of $N$ and $P$ is suppressed to a visual-acceptable level.
\textbf{(3) color neutrality}: The gray levels calculated from the RGB channels should be consistent~\citeps{buchsbaum1980spatial}.
However, traditional cameras sometimes fail to capture sufficient details in extreme-lighting scenes.
Under such low-light conditions, images may lack visible details and be contaminated by noise, while in high-light conditions, images may suffer from oversaturation, losing texture and edge information.

Event cameras asynchronously detect illumination changes at each pixel, making them ideal for capturing scenes with extreme or rapidly changing lighting conditions~\citeps{gallego2020event}. 
The event stream's outputs are formatted as 4 components: $(x, y)$ (pixel coordinates), $t$ (timestamp), and $p \in \{+1, -1\}$ (polarity, indicating light intensity increase or decrease). 
Events are triggered when the change in illumination exceeds a threshold $C$ ($\Delta L = \log(L(t)) - \log(L(t-\Delta t))$ where $|\Delta L| > C$).
We jointly leverage the complementary information from an image $I_{rgb}$ and its corresponding events $E$ to recover a high-quality well-illuminated image $\hat{I_{\text{rgb}}}$ that accurately represents the scene radiance $L(t)$, while also allowing for adjustable brightness. 
To achieve this, we introduce a brightness prompt $B$ that controls the overall brightness of the output image.
This allows us to map the $L(t)$ into an image that is optimally exposed for human observation.
Our task setting can thus be formulated as Eq.~\ref{eq:our_taget}, where $f_{see}$ is our proposed model.
\begin{equation}
f_{see}(I_{rgb}, E, B) \rightarrow \hat{I_{\text{rgb}}}.
\label{eq:our_taget}    
\end{equation}
This formulation has two advantages: \textbf{(1) robust training}: By inserting the brightness prompt $B$ during training, we can decouple the model from biases in the training data with specific brightness levels, enabling the model to generalize better over illuminates domain.
\textbf{(2) flexible inference}: During inference, the prompt $B$ can be set to a default value (\eg, $B = 0.5$) to produce images with general brightness, or be adjusted to achieve different brightness levels, providing flexibility for applications requiring specific exposure adjustments or artistic effects.
\textit{Please refer to the appendix for more details of this section.}

\noindent\textbf{Differences with HDR Image Reconstruction:}
In this section, we define the brightness adjustment task based on events. 
It is important to note that HDR image reconstruction also involves processing both low-light and high-light images to capture more details in both dark and bright regions. 
However, HDR image reconstruction aims to extend the dynamic range, which is a more ambitious and challenging task. 
In contrast, our goal is to adjust the dynamic range, which is easier to implement and more flexible. 
We highlight three key differences between these two tasks:
\textbf{(1)} \textit{Different Objectives:}
HDR image reconstruction aims to expand the dynamic range of an image by combining multiple exposures, capturing details in both dark and bright regions. Mathematically, this involves recovering a radiance map \( R(x) \) from events and images captured at different exposure levels, represented as \( R(x) = f^{-1}(I_{\text{LDR}}(x),E) \), where \( I_{\text{LDR}}(x) \) is the observed LDR image. In contrast, our brightness adjustment task focuses on modifying the exposure of a single image to recover lost details, without necessarily expanding the dynamic range, as shown in Eq.~\ref{eq:our_taget}.
\textbf{(2)} \textit{Distinct Challenges:}
HDR image reconstruction seeks to transfer the HDR characteristics of events to LDR images, requiring an HDR image as supervision. In contrast, our brightness adjustment task uses event data to adjust the exposure of a single image, without needing multiple images at different exposures for supervision.
\textbf{(3)} \textit{Different Dataset Construction:}
HDR reconstruction datasets require capturing multiple images at different exposure levels~\citep{messikommer2022multi}, which can be computationally expensive and difficult to handle in dynamic scenes. In contrast, our approach uses event cameras to capture pairs of images and events under varying lighting conditions, simplifying data construction and making it more adaptable to dynamic scenes.

\begin{figure*}[htbp!]
\centering
\includegraphics[width=\linewidth]{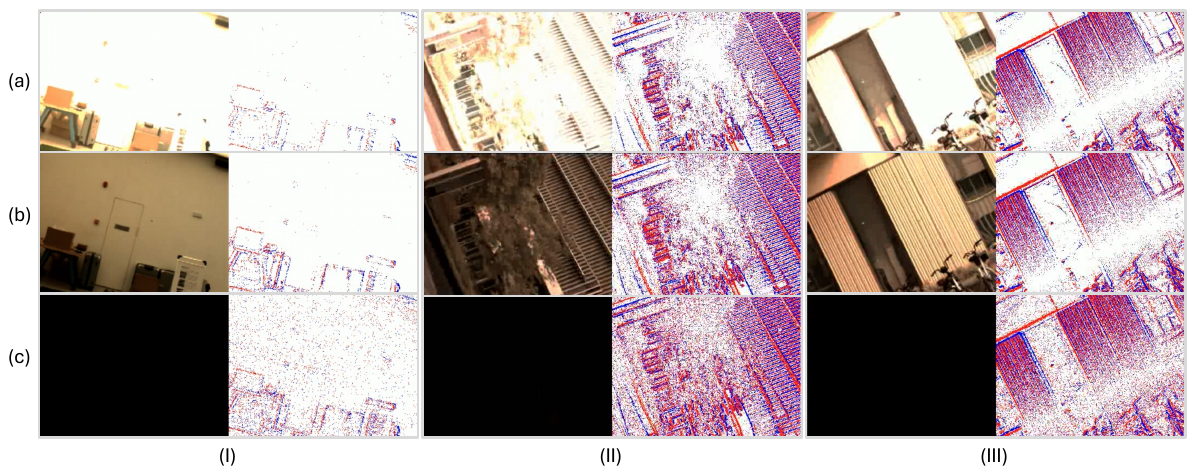}
\caption{Visualization of different lighting conditions in SEE-600K. We present three scenes captured in SEE-600K, demonstrating its broad illumination coverage. Each row corresponds to a different lighting condition within the same scene: (a) High-Light, (b) Normal-Light, and (c) Low-Light. The columns (I), (II), and (III) represent different environments. Each sub-image pair consists of a frame (left) and its corresponding events (right), highlighting the ability of event cameras to capture fine-grained temporal changes across varying lighting conditions.
\label{fig:11-Dataset-Sample}}
\end{figure*}
\begin{figure*}[t!]
\centering
\includegraphics[width=1\linewidth]{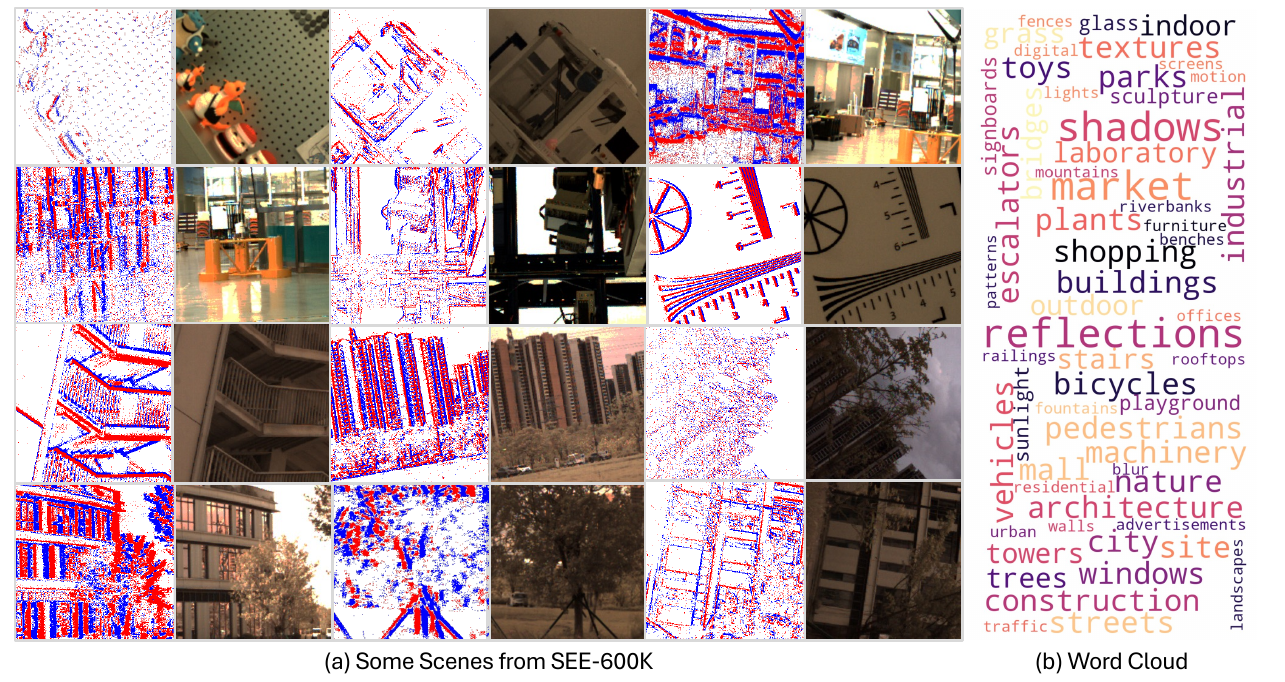}
\caption{Visualization of scene diversity in SEE-600K. The left side (a) presents 12 representative scenes from SEE-600K, showcasing the variety of environments captured in our dataset, including urban areas, buildings, natural scenes, industrial settings, and structured indoor spaces. With a total of 202 distinct scenes, SEE-600K provides a broad range of lighting conditions and structural diversity for event-based imaging research. The right side (b) displays a word cloud generated from our dataset, illustrating the richness of scene categories and objects present in SEE-600K, further emphasizing its comprehensiveness.}
\label{fig:12-MoreExamples-WordCloud}
\end{figure*}
\begin{table*}[t!]
\centering
\caption{Comparison of different datasets in event-based imaging. Our SEE-600K dataset captures a broader range of illumination conditions, encompassing low-light, normal-light, and high-light settings. Additionally, it uniquely provides multiple groups of recordings per scene, enabling more robust learning of brightness transitions across various lighting conditions. This comprehensive dataset offers significantly larger scale and diversity, making it a valuable resource for event-based imaging research.}
\renewcommand{\arraystretch}{1.2} %
\setlength{\tabcolsep}{4pt} %
\resizebox{\linewidth}{!}{
\begin{tabular}{lccr|p{1.8cm}|cccc}
\toprule
\multirow{2}{*}{\textbf{Dataset}} &
\multirow{2}{*}{\makecell[c]{\textbf{Public}\\\textbf{Available}}} &
\multirow{2}{*}{\makecell[c]{\textbf{Dynamic}\\\textbf{Motion}}} &
\multirow{2}{*}{\makecell[c]{\textbf{Size}\\(\textit{Frames})}} &
\multirow{2}{*}{\makecell[c]{\textbf{Device}}} &
\multicolumn{4}{c}{\textbf{Frames}} \\
\cmidrule(lr){6-9}
& & & & &
\textbf{Low-Light} &
\textbf{Normal-Light} &
\textbf{High-Light} &
\textbf{Multiple Groups} \\
\midrule
DVS-Dark~\citep{zhang2020learning}
& \xmark    & \cmark & 17,765           & DAVIS240C       & \cmark & \xmark & \xmark & \xmark \\
LIE~\citep{jiang2023event}
& \cmark    & \xmark & 2,231            & DAVIS346        & \cmark & \cmark & \xmark & \xmark\\
EvLowLight~\citep{liang2023coherent}
& \xmark    & \cmark & \textemdash{}    & DAVIS346        & \cmark & \cmark & \xmark & \xmark\\
RLED~\citep{liu2024seeing}
& \xmark    & \cmark & 64,200           & Prophesee~EVK4  & \xmark & \cmark & \xmark & \xmark\\
SDE~\citep{liang2024towards}
& \cmark    & \cmark & 31,477           & DAVIS346        & \cmark & \cmark & \xmark & \xmark\\
SEE-600K (\textbf{Ours})
& \cmark    & \cmark & \textbf{610,126} & DAVIS346        & \cmark & \cmark & \cmark & \cmark\\
\bottomrule
\end{tabular}
}
\label{tab:event_datasets}
\end{table*}

\section{SEE-600K Dataset\label{sec:see_dataset}}

Based on the new problem defined in the previous section, we require a large-scale dataset to support training and evaluation, which includes events and images under various lighting conditions, with complex motion trajectories and spatio-temporal alignment. 
To address this, we propose a new dataset, SEE-600K. 
In this section, we introduce the SEE-600K dataset, focusing on two aspects: the dataset collection method and its diversity.

\subsection{Collection Method}

To achieve this collection, we propose the following solution, which consists of three key components.
\noindent\textbf{(1) multiple lighting conditions}:
Our approach is based on the principle that lighting transitions continuously from low to high intensity. 
Unlike previous datasets~\citeps{liang2024towards,wang2021seeing}, which captured only a \textit{single} pair of low-light and normal-light conditions, we focus on \textit{multiple} samples, as shown in Fig.~ \ref{fig:11-Dataset-Sample}.
To cover a broader lighting range, we record an average of four videos per scene, using neutral density (ND) filters at three levels $(1/8, 1/64, 1/1000)$ and one without a filter. 
We also adjust the aperture and exposure settings to capture each scene under diverse lighting conditions.
\textbf{(2) complex motion trajectories}:
We employ the Universal Robots UR5e robotic arm, which can provide high stability and repeat the same non-linear trajectory with an error margin of \textbf{0.03 mm}~\citeps{liang2024towards,brey2024smartphone}, allowing us to capture multiple videos with spatial consistency, as exhibited in Fig.~\ref{fig:2-Reg-Vis-Release} (a).
\textbf{(3) spatio-temporal alignment}:
While the robotic arm guaranteed spatial alignment, asynchronous control over the camera's start and stop times inevitably introduced timing deviations.
To resolve this, we propose an IMU-based temporal alignment algorithm, as shown in Fig.~\ref{fig:2-Reg-Vis-Release} (b).
IMU streams synchronized to events and videos with microsecond timestamps in the DVS346 camera.
Additionally, the IMU stream depends only on motion trajectory and enjoys a temporal resolution of $1000$ Hz.
Based on this, our algorithm achieves precise temporal alignment, ensuring synchronization across the entire dataset, as displayed in Fig.~\ref{fig:2-Reg-Vis-Release} (c).
We introduce the IMU-based alignment method, and the alignment evaluation results as follows.

\begin{figure*}
\centering
\includegraphics[width=1\linewidth]{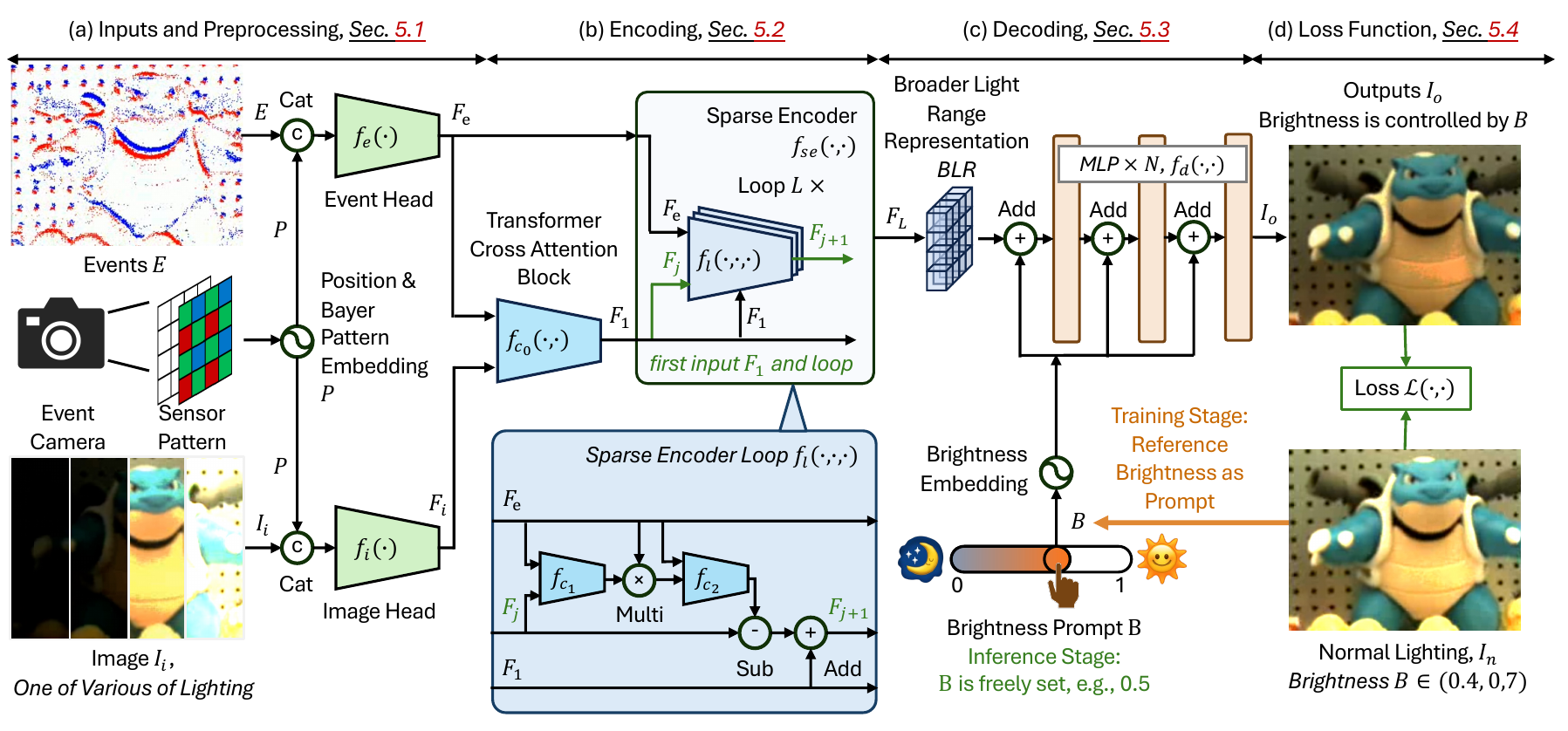}
\caption{{Overview of our proposed framework, called SEE-Net, which is composed of four stages: \textbf{(a)} Inputs and Preprocessing, \textbf{(b)} Encoding, \textbf{(c)} Decoding, and \textbf{(d)} Loss Function. This framework takes as input an image captured under a wide range of lighting conditions, along with its corresponding events. The output is a brightness-controllable image, where the brightness is guided by the brightness prompt $B$, enabling flexible pixel-level adjustment during inference.}}
\label{fig:5-SEENet-Framework-Release}
\end{figure*}

\noindent\textbf{Temporal IMU Registration Algorithm\label{sec:temp-imu-reg}:}
We propose an IMU data registration algorithm that aligns the source sequence $S$ and target sequence $T$ by finding the optimal bias $b$ and matching length 
$l$ to minimize the $L_1$ distance between them.
Given the high resolution of IMU data at $1000 Hz$, an exhaustive search for the optimal bias is computationally infeasible. 
To address this, we introduce a multi-level iterative strategy.
First, we denoise the IMU data using a Kalman filter~\citeps{mirzaei2008kalman}.
Then, the average pooling is utilized to reduce the sequences to two additional levels, Level-1 ($S_1$, $T_1$) and Level-2 ($S_2$, $T_2$), as shown in Fig.~\ref{fig:2-Reg-Vis-Release}~(a)-{\textcircled{1}\textcircled{2}}.
This reduces computational complexity while preserving essential alignment features. The window size is chosen based on our video durations, which range from 10 to 120 seconds.
We perform a coarse search for the optimal bias $b$ and matching length $l$ at the lowest resolution (Level-2).
The results from this level serve as center points for finer searches at higher resolutions.
Specifically, the bias and length identified at each level guide local searches at the next level up, as displayed in Fig.~\ref{fig:2-Reg-Vis-Release}~(a)-{\textcircled{3}\textcircled{4}\textcircled{5}}.
At Level-1 and the original data level (Level-0), we only need to search locally around these center points.
This hierarchical approach efficiently achieves high matching accuracy with significantly reduced computational effort.

\noindent\textbf{Spatial-Temporal Alignment Evaluation:}
To evaluate the accuracy of our IMU registration algorithm, we capture the same scene twice under identical lighting conditions, as illustrated in Fig.~\ref{fig:2-Reg-Vis-Release}~(b).
We assess the alignment metric between the two image sequences by calculating the pixel-level distance at the corresponding timestamp.
\textit{\textbf{Alignment Metric}}: For each image pair, we extract keypoints using SIFT~\citeps{lowe2004distinctive} and then employ the FLANN matcher~\citeps{muja2009fast} to find matching keypoints between the two images. 
Based on these matched keypoints, we compute the affine transformation matrix using RANSAC~\citeps{fischler1981random}.
This transformation is subsequently applied to each pixel, allowing us to calculate the displacement distance for every pixel.
Finally, the average pixel distance is employed as the metric for alignment.
\textit{\textbf{Alignment Results}}: In the alignment evaluation, we select scenes with well-defined textures, as illustrated in Fig.~\ref{fig:2-Reg-Vis-Release} (b).
After calculating the pixel distances, we observe that the average pixel error between the paired images is $0.2967$ pixels.
Throughout the entire time sequence, the pixel-level distance remains below $0.8$ pixels, with the majority of errors being under $0.5$ pixels, as exhibited in Fig.~\ref{fig:2-Reg-Vis-Release} (c).
These results demonstrate that the registration accuracy of our dataset reaches sub-pixel precision.

\subsection{Diversity Analysis}
To analyze the diversity of our dataset, we evaluate the SEE-600K dataset in terms of both illumination and scene variety.
Firstly, as shown in Fig.~ \ref{fig:11-Dataset-Sample}, for each scene, we capture multiple lighting conditions, including high-light, normal-light, and low-light, ensuring a broad range of illumination coverage. This approach enhances the dataset's ability to support tasks requiring robust learning across diverse lighting variations.
Secondly, the SEE-600K dataset includes 202 distinct scenes, as shown in Fig.~ \ref{fig:12-MoreExamples-WordCloud}, with a total of over 600K data samples, making it approximately 20 times larger than the latest SDE dataset~\citep{liang2024towards}. 
This substantial increase in scale provides a richer foundation for training models with improved generalization capabilities.
Additionally, we compare SEE-600K with several other event-based datasets, as shown in Tab.~ \ref{tab:event_datasets}.
Unlike previous datasets such as DVS-Dark~\citep{zhang2020learning} or EvLowLight~\citep{liang2023coherent}, SEE-600K includes a broader range of illumination conditions.
Furthermore, SEE-600K uniquely offers multiple groups of recordings per scene, enabling more robust learning of brightness transitions across various lighting conditions.
This makes SEE-600K a significantly more comprehensive resource in terms of scale and diversity compared to other datasets like LIE~\citep{jiang2023event}, RLED~\citep{liu2024seeing}, and SDE~\citep{liang2024towards}, which either lack diversity in lighting conditions or include fewer frames.
Finally, we assess the diversity of the dataset using a word cloud generated by ChatGPT-4o \citep{hernandez2025assessment}, as shown in Fig.~\ref{fig:12-MoreExamples-WordCloud}~(b).
This word cloud illustrates the variety of scene categories and objects present in the dataset, further emphasizing its comprehensive coverage.
With scenes ranging from urban environments to industrial settings and natural landscapes, SEE-600K captures a diverse array of objects and contexts, making it a valuable resource for event-based imaging research.

\section{Methods\label{sec:method}}
\textbf{Overview:} As shown in Fig.~\ref{fig:5-SEENet-Framework-Release}, our framework, SEE-Net, consists of four implementation parts: \textbf{(a)} Inputs and Preprocessing, \textbf{(b)} Encoding, \textbf{(c)} Decoding, and \textbf{(d)} Loss Function.
The input is an image $I_i$ and its corresponding events $E$. 
The output is a brightness-adjustable image $I_o$, where the brightness is controlled by the prompt $B \in (0,1)$.
During training, the brightness prompt $B$ is calculated according to the target image.
On the other hand, during testing, $B$ can be freely set, with a default value of 0.5, which follows the exposure control constraint \citeps{mertens2009exposure,mertens2007exposure}. 
Overall, the SEE-Net $f_{see}$ can be described by the Eq.~\ref{eq:f_see} to match our learning task in Sec.~\ref{sec:preliminary}.
\begin{equation}
I_o = f_{\text{see}}(I_i, E, B).
\label{eq:f_see}
\end{equation}
Below, we elaborate the insights and implementation details of each part.

\noindent\textbf{Inputs and Processing:}
This part aims to transform initial inputs into features that retain original information for the encoding stage.
The inputs consist of the image $I_i$ and the events $E$, where $I_i$ has a dimension of $H \times W \times 3$ (with $H$ and $W$ representing the height and width, and 3 representing the color channel number). 
The event stream $E$ is represented as a voxel grid~\citeps{tulyakov2022time} with a dimension of $H \times W \times M$, where $M$ represents the number of time slices of events.
The events include color information~\cite{scheerlinck2019ced}, which was overlooked in previous works, \eg, \citeps{liang2024towards,liang2023coherent}.
Specifically, this DVS346 sensor records events with Bayer Pattern~\citeps{lukac2005color}.
To effectively embed both the color and positional information during framework training, we design the position and Bayer Pattern embeddings, as shown in Fig.~\ref{fig:5-SEENet-Framework-Release} (a).
The position and Bayer Pattern are denoted as a vector $(x, y, bp)$, where $x,y$ is the pixel position, and $bp$ denotes the Bayer Pattern index, which takes a value from 0 to 3.
We embed this vector into a higher-dimensional feature, termed as $P$, and concatenate it with the inputs.
Two layers $1\times 1$ convolutions, denote $f_e$ and $f_i$, are then applied to obtain the initial event features $F_e$ and image features $F_i$.
This process is described by the Eq.~\ref{eq:f_head}, where $f_{cat}$ denotes the concatenation function.
\begin{equation}
\begin{aligned}
F_e &= f_e(f_{cat}(E, P)), \\
F_i &= f_i(f_{cat}(I_i, P)). 
\end{aligned}
\label{eq:f_head}
\end{equation}

\begin{table*}[t!]
\centering
\caption{{Comparison of different methods on the SDE dataset. 
The best performances is highlighted in \textbf{bold}.
\rrev{$\dag$ refers to the original model for the HDR task, which is fine-tuned and trained on SDE}
}}
\resizebox{\linewidth}{!}{
\setlength{\tabcolsep}{0.018\linewidth}{
\begin{tabular}{lrrclllllll}
\toprule
\multirow{2}{*}{\textbf{Method}} & 
\multirow{2}{*}{\textbf{FLOPs}} & 
\multirow{2}{*}{\textbf{Params}} & 
\multirow{2}{*}{\textbf{Events}} & 
\multicolumn{2}{c}{\textbf{indoor}} & 
\multicolumn{2}{c}{\textbf{outdoor}} & 
\multicolumn{2}{c}{\textbf{average}} \\ 
\cmidrule(lr){5-6} \cmidrule(lr){7-8} \cmidrule(lr){9-10}
 &  &  &  & PSNR   & SSIM   & PSNR   & SSIM   & PSNR   & SSIM  \\ 
\midrule
DCE~\citeps{guo2020zero}           
& 0.66      & 0.01        & \XSolidBrush      & 13.91  & 0.2659 & 13.38   & 0.1842  & 13.64  & 0.2250  \\
SNR~\citeps{xu2022snr}           
& 26.35     & 4.01        & \XSolidBrush      & 20.05  & 0.6302 & 22.18   & 0.6611  & 21.12  & 0.6457  \\ 
UFormer~\citeps{wang2022uformer}       
& 12.00     & 5.29        & \XSolidBrush      & 21.09  & 0.7524 & 22.32   & 0.7469  & 21.71  & 0.7497  \\ 
LLFlow~\citeps{wu2023learning}
& 409.50    & 39.91       & \XSolidBrush      & 20.92  & 0.6610 & 21.68   & 0.6467  & 21.30  & 0.6539  \\ 
Retinexformer~\citeps{cai2023retinexformer}
& 15.57     & 1.61        & \XSolidBrush      & 21.30  & 0.6920 & 22.92   & 0.6834  & 22.11  & 0.6877  \\ 
E2VID+~\citeps{stoffregen2020reducing}        
& 27.99     & 10.71       & \Checkmark        & 15.19  & 0.5891 & 15.01   & 0.5765  & 15.10  & 0.5828  \\ 
ELIE~\citeps{jiang2023event}          & 440.32    & 33.36       & \Checkmark        & 19.98  & 0.6168 & 20.69   & 0.6533  & 20.34  & 0.6350  \\ 
\rrev{HDRev\cites{yang2023learning} $\dag$}
&   \rrev{118.65}    & \rrev{13.42} & \rrev{\Checkmark}  & \rrev{21.13}    & \rrev{0.6239}      & \rrev{21.82}   & \rrev{0.6824}    &  \rrev{21.47} & \rrev{0.6531}   \\
\rrev{\cites{wang2023event}}    &  \rrev{170.32}     
&   \rrev{7.38}      & \rrev{\Checkmark}        & \rrev{21.29}   & \rrev{0.6786}  & \rrev{22.08}   & \rrev{0.7052}  & \rrev{21.68}  & \rrev{0.6919}  \\ 
eSL-Net~\citeps{wang2020event}       & 560.94    & 0.56        & \Checkmark        & 21.25  & 0.7277 & 22.42   & 0.7187  & 21.84  & 0.7232  \\ 
\cites{liu2023low}    & 44.71     & 47.06       & \Checkmark        & 21.79  & 0.7051 & 23.35   & 0.6895  & 22.57  & 0.6973  \\ 
EvLowlight~\citeps{liang2023coherent}    & 524.95    & 15.49       & \Checkmark        & 20.57  & 0.6217 & 20.04   & 0.6485  & 20.31  & 0.6351  \\ 
EvLight~\citeps{liang2024towards}       & 180.90    & 22.73       & \Checkmark        & \textit{22.44}  & \textit{0.7697} & \textit{23.21}   & \textit{0.7505}  & \textit{22.83}  & \textit{0.7601}  \\
SEENet (Ours) & 405.72    & 1.90        & \Checkmark        & \textbf{22.54}  & \textbf{0.7756} & \textbf{24.60}   & \textbf{0.7692}  & \textbf{23.57}  & \textbf{0.7724}  \\ 
\bottomrule
\end{tabular}
}
}
\label{tab:sde_dataset}
\end{table*}

\begin{figure*}
\centering
\includegraphics[width=0.94\linewidth]{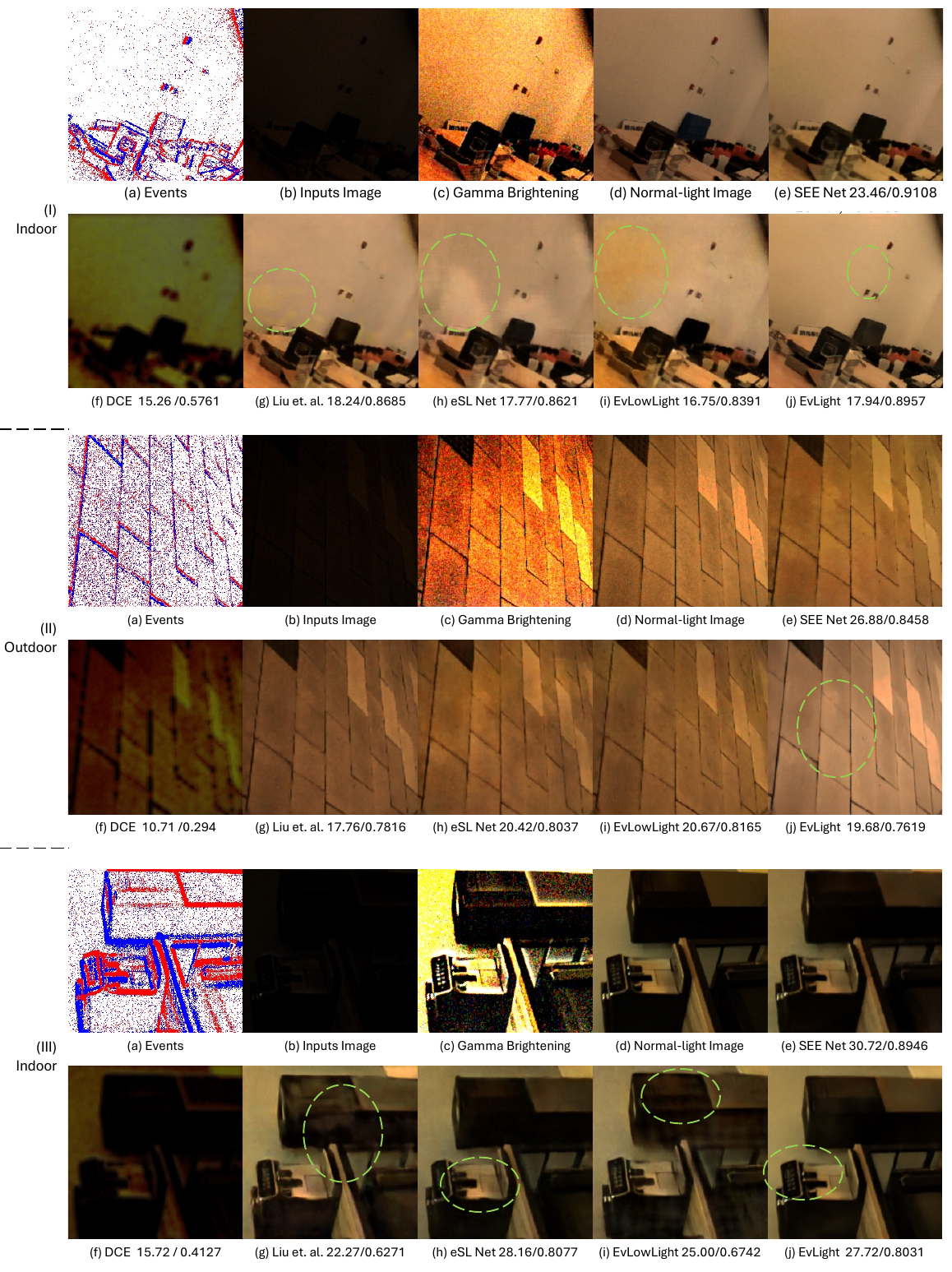}
\caption{{Visualization results of three scenes from the SDE dataset under varying lighting conditions.}}
\label{fig:6-SDE-VisualizationResults-Release}
\end{figure*}

\textbf{Encoding:}
In this stage, we aim to obtain the BLR by employing the event feature $F_e$ to enhance the image feature $F_i$, facilitating noise reduction and the acquisition of broader light range information.
Since $F_e$ contains rich information about the lighting changes across different intensity levels, we use it as the source for representing the broader light range.
However, event data only records changes in illumination, which differ fundamentally from the static RGB frame modality. This makes directly utilizing event data for broader light representation challenging.
To address this, we employ a cross-attention~\citeps{liang2021swinir} for feature fusion, producing the initial fused broad-spectrum feature $F_1$, expressed as $F_1 = f_{c_0}(F_e, F_i)$, where $f_{c_0}$ is a cross-attention block.
Then, inspired by previous works~\citeps{wang2020event}, we utilize sparse learning to generate residuals for $F_1$ from the event features $F_e$.
These residuals are progressively generated from the loop that executes $L$ times.
Multiple iterations are used because they allow the model to iteratively refine the residuals, capturing finer details and enhancing the feature representations by progressively integrating information from the events.
A single loop of this process can be expressed as, $F_{j+1} = f_l(F_e, F_1, F_j)$, where $f_l$ is a loop function that contains two cross-attention blocks as shown in Fig.~\ref{fig:5-SEENet-Framework-Release} (b), where $F_j$ and $F_{j+1}$ are the input and output of one loop.
After $L$ iterations, the final feature $F_L$ represents the BLR, as described by Eq.~\ref{eq:f_se}.
\begin{equation}
\begin{aligned}
F_L &= f_{se}(F_e, F_1) \\
    &= f_l(F_e, F_1, f_l(F_e, F_1, ... f_l(F_e, F_1, F_1))) \\
    &= f_l \left( F_e, F_1, \underbrace{f_l\left( F_e, F_1, \dots, f_l\left( F_e, F_1, F_1 \right) \right)}_{\text{recursive part}} \right)
\end{aligned}
\label{eq:f_se}
\end{equation}
\textbf{Decoding:} The objective of this part is to decode the BLR into a brightness-adjustable image $I_o$. 
In designing this decoder, we focus on two key insights: 
(1) The decoding process should be pixel-wise and efficient, allowing for greater flexibility during model deployment;
(2) The embedding of the brightness information should be thorough and fully integrated.
With these insights, we design the decoder with only a 5-layer MLP as shown in Fig~\ref{fig:5-SEENet-Framework-Release} (c).
Our decoder begins by encoding the brightness prompt $B \in (0,1)$ into an embedding vector.
To effectively encode the high-frequency brightness prompt into features that are easier for the network to learn~\citeps{vaswani2017attention}, we introduce a learnable embedding, denoted as $\vvb{B}=f_{pe}(B)=f_{mlp}(f_{cat}(f_{mlp}(B),B))$, which consists of two MLP layers.
Through this embedding, the brightness prompt $B$ is transformed into a vector $\vvb{B}$, matching the dimensions of the BLR channels.
We then integrate this embedding $\vvb{B}$ into the decoder.
To ensure the brightness prompt is fully incorporated and prevent information loss through multiple MLP layers, we employ a multi-step embedding approach, as displayed in Eq.~\ref{eq:f_d}, which guarantees that the brightness is progressively embedded throughout the decoding process.
During the training phase, the prompt $B$ is derived from the reference image by applying $f_b$ to calculate the global average brightness.
In contrast, during the testing phase, $B$ can be set freely, with a typical example being a value of 0.5.
\begin{equation}
\begin{aligned}
I_o &= f_d(F_L,\vvb{B}) \\
    &= f_{mlp}(\vvb{B}+f_{mlp}(\vvb{B}+...f_{mlp}(\vvb{B} + F_L))) \\
    &= f_{mlp} \left( \vvb{B} + \underbrace{f_{mlp} \left( \vvb{B} + \dots + \underbrace{f_{mlp} \left( \vvb{B} + F_L \right)}_{\text{first layer}} \right)}_{\text{recursive part}} \right).
\end{aligned}
\label{eq:f_d}
\end{equation}

\noindent\textbf{Loss Function:} 
The purpose of our loss function is to supervise the prediction $I_o$ using the ground truth $I_t$, with the corresponding brightness $B=f_b(I_t)$. The loss function consists of two main components: image reconstruction loss $\mathcal{L}_{i}$ and gradient loss $\mathcal{L}_{g}$. 
The image reconstruction loss is Charbonnier loss~\citeps{lai2018fast}, which effectively handles both small and large errors. 
Additionally, we employ gradient loss to improve the structural consistency of the output image.
This is achieved by enforcing $L_1$ constraints on the gradients of both the output and ground truth images. 
Therefore, the overall loss function is formulated as a weighted sum of the image loss and gradient loss, as exhibited in Eq.~\ref{eq:loss}.
Here, $\nabla$ denotes the gradient operator, and $\lambda_1$ and $\lambda_2$ are the weights that balance the contributions of two loss terms.
\begin{equation}
\begin{aligned}
\mathcal{L}(I_o,I_t) &= \lambda_1 \mathcal{L}_{i} + \lambda_2 \mathcal{L}_{g} \\
&= \lambda_1  \sqrt{(I_o  - I_t)^2 + \epsilon^2} +  \lambda_2 \Vert \nabla I_o  - \nabla I_t \Vert.
\end{aligned}
\label{eq:loss}
\end{equation}

\section{Experiments\label{sec:experiments}}

\begin{table*}[t!]
\centering
\caption{Evaluation on the SEE-600K dataset, with methods trained on both the SDE and SEE-600k. 
}
\resizebox{\linewidth}{!}{
\setlength{\tabcolsep}{0.011\linewidth}{
\begin{tabular}{llrrrrrrrrr}
\toprule
\multirow{2}{*}{\makecell[c]{\textbf{Training}\\ \textbf{Dataset}}} 
& \multirow{2}{*}{\textbf{Methods}} 
& \multicolumn{3}{c}{\textbf{low light}} 
& \multicolumn{3}{c}{\textbf{high light}} 
& \multicolumn{3}{c}{\textbf{normal light}} \\
\cmidrule(lr){3-5} \cmidrule(lr){6-8} \cmidrule(lr){9-11}
& 
& \makecell[c]{PSNR}  & \makecell[c]{SSIM}   & \makecell[c]{$L_1$}   & \makecell[c]{PSNR}  & \makecell[c]{SSIM}   & \makecell[c]{$L_1$}   & \makecell[c]{PSNR}   & \makecell[c]{SSIM}   & \makecell[c]{$L_1$}   \\
\midrule
\multirow{6}{*}{SDE}
& DCE \citep{guo2020zero}            
& 9.10       & 0.0968       & 0.3572     & 6.26      &  0.3419       & 0.4649     & \textbf{10.79}       & \textbf{0.3992}       & \textbf{0.2524}     \\
& eSL Net~\citep{wang2020event}        
& 11.92      & 0.3275       & 0.2703     & \textbf{6.66}      & \textbf{0.1672}       & \textbf{0.4001}     & 7.65       & 0.2685       & 0.3481     \\
& \cite{liu2023low}     
& 12.41      & 0.4001       & 0.2487     & 5.53      & 0.1950       & 0.4534     & 6.58       & 0.2805       & 0.4129     \\
& EvLowLight \citep{liang2023coherent}     
& 12.68      & 0.4341       &  0.2338    & 4.11      & 0.3071       & 0.6062     & 7.01       & 0.3950       & 0.4520     \\
& EvLight \citep{liang2024towards}       
& 13.07      & 0.4651       & 0.2337     & 5.12      & 0.1005       & 0.4842     & 6.29       & 0.2805       & 0.4336     \\
& SEENet         & \textbf{14.84}      & \textbf{0.5693}       & \textbf{0.1779}     & 3.84    & 0.2119       & 0.6123     & 5.36       & 0.2980       & 0.5056     \\
\midrule
\multirow{5}{*}{SEE}
& eSL Net~\citep{wang2020event} 
& 11.95      & 0.3845       & 0.2421     & 12.84      & 0.4660       & 0.2076     & 13.45       & 0.5682       & 0.1957     \\
& {EvLowLight} \citep{liang2023coherent}
& \rrev{12.83}      & \rrev{0.4511}       & \rrev{0.2151}     & \rrev{12.79}      & \rrev{0.4696}       & \rrev{0.2084}     & \rrev{13.04}       & \rrev{0.5531}       & \rrev{0.2144} \\
& \cite{liu2023low} 
& 13.48      & 0.5068       & 0.1946     & 12.30      & 0.4766       & 0.2221     & 13.70       & 0.5474       & 0.2151     \\
& EvLight \citep{liang2024towards}
& 13.70      & 0.5150       & 0.1960     & 13.45      & 0.4918       & 0.1990     & 13.63       & 0.5924       & 0.2004     \\
& SEENet         & \textbf{18.77}      & \textbf{0.6303}       & \textbf{0.0971}     & \textbf{19.21}      & \textbf{0.6675}       & \textbf{0.0806}     & \textbf{20.92}       & \textbf{0.8002}       & \textbf{0.0606}     \\
\bottomrule
\end{tabular}
}
}
\label{tab:see_comp}
\end{table*}

\begin{figure*}
\centering
\includegraphics[width=1\linewidth]{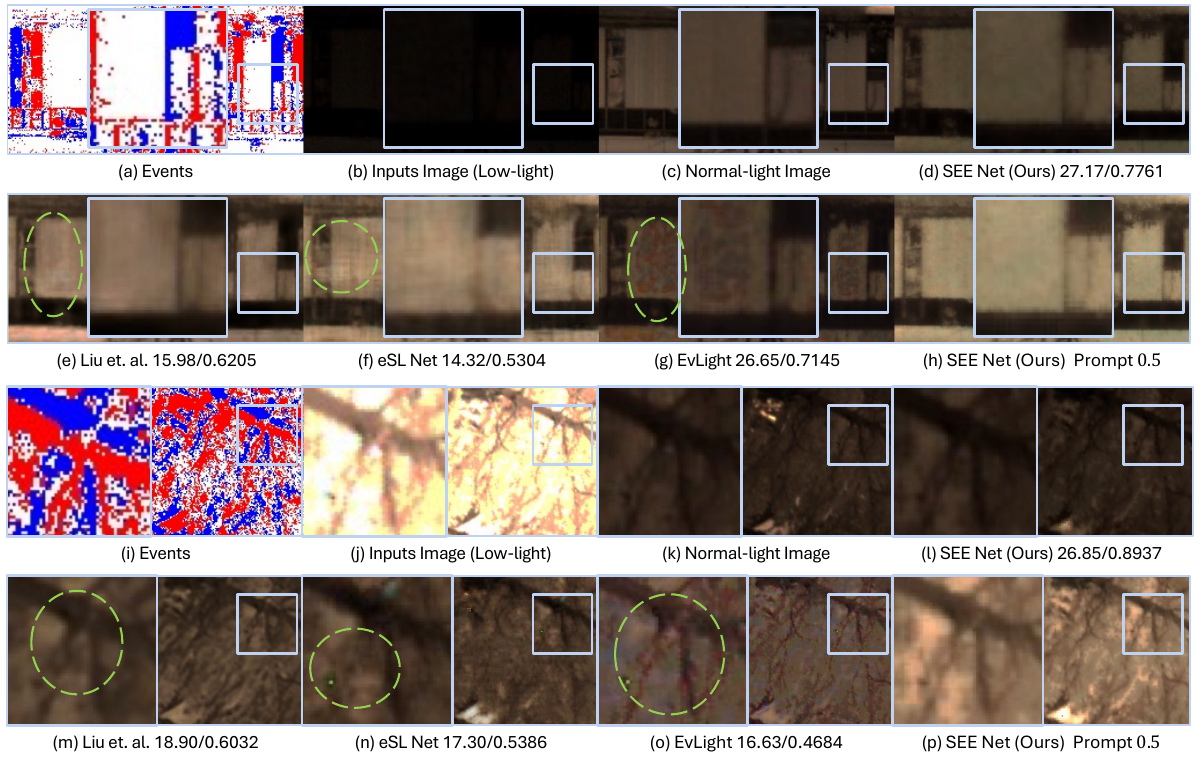}
\caption{{Visual examples of low-light enhancement and high-light recovery on the SEE-600K dataset.}}
\label{fig:7-SEE-Dataset-Release}
\end{figure*} 

\begin{figure*}
\centering
\includegraphics[width=0.84\linewidth]{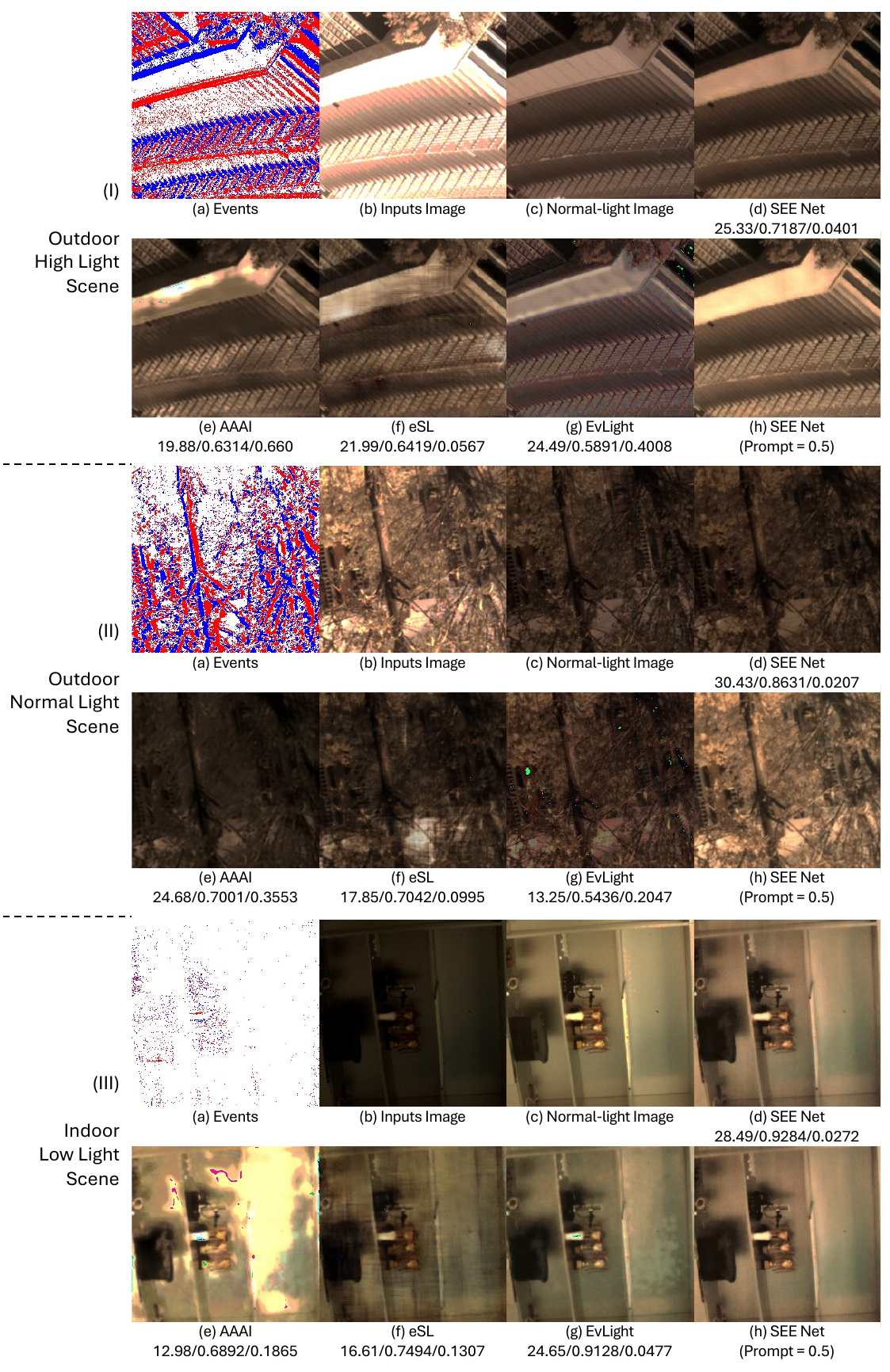}
\caption{{Visual examples of low-light enhancement and high-light recovery on the SEE-600K dataset.}}
\label{fig:14-SEE-Dataset-More}
\end{figure*}

\begin{figure*}
\centering
\includegraphics[width=1\linewidth]{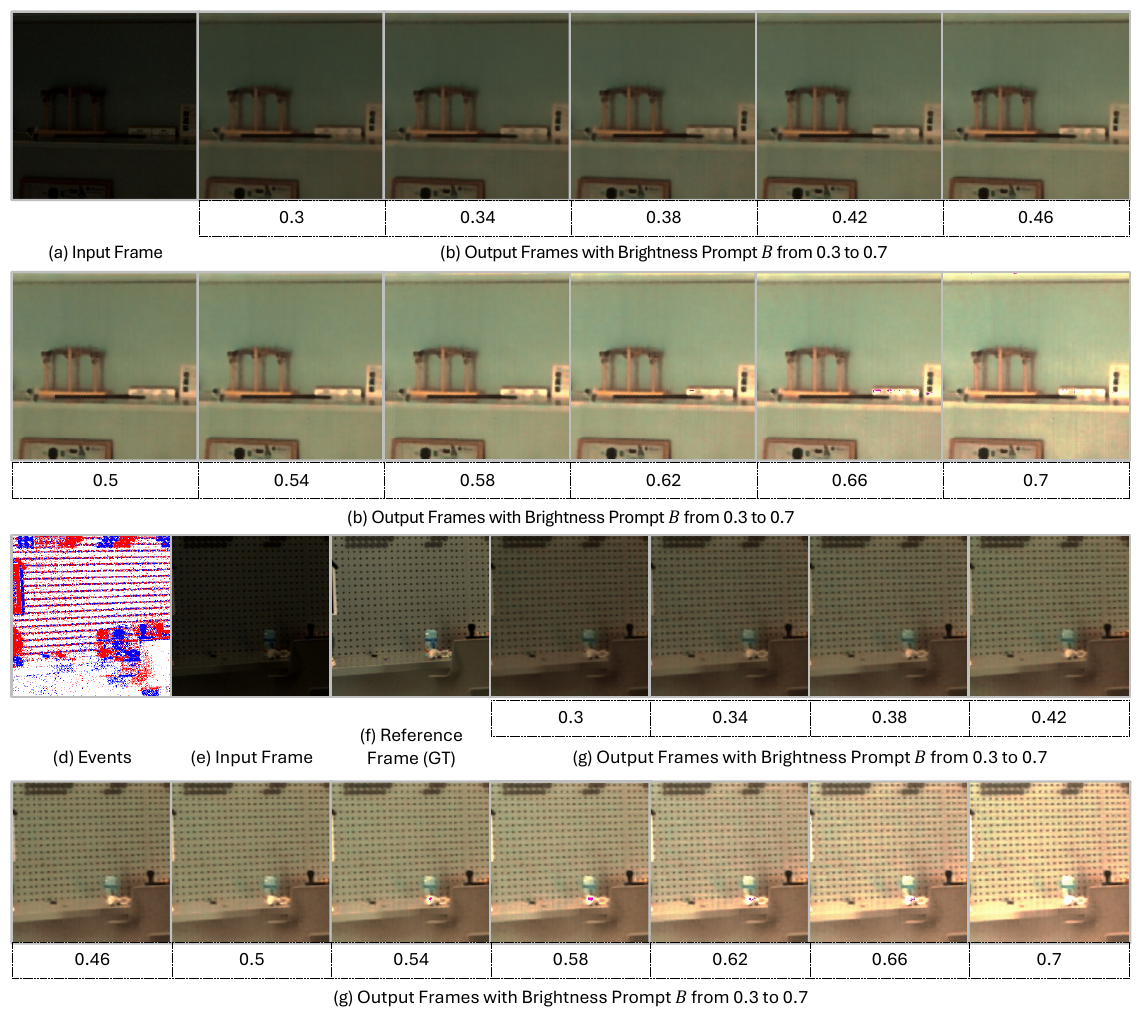}
\caption{ Visualization of brightness adjustment using varying brightness prompts $B$ from 0.3 to 0.7, showing smooth brightness transitions in SEE-600K dataset.\rrev{For more visualizations, see the Appendix.}}
\label{fig:8-Prompt0.3to0.7}
\end{figure*}

\noindent\textbf{{Implementation Details:}}
Our experiments use the Adam optimizer with an initial learning rate of $2e-4$ for all the experiments.
We train our model for 40 epochs on the SDE dataset~\citeps{liang2024towards}.
On the SEE-600K dataset, we train for only 20 epochs, as SEE-600K is extremely large. 
All of our training is conducted on an HPC cluster, with a batch size of 2.
To enhance data diversity, we apply random cropping to the images and perform random flips and rotations.

\noindent\textbf{{Evaluation Metrics:}} We maintain consistency with previous methods~\citeps{liang2024towards,liang2023coherent} by using PSNR and SSIM~\citeps{wang2004image}. 
However, since our proposed new problem is highly challenging and most current approaches perform poorly on our SEE-600K dataset, we additionally introduce the $L_1$ distance as a reference.

\noindent\textbf{Dataset:} We conduct experiments on two real-world datasets: 
\textbf{(1) SDE}~\citeps{liang2024towards} comprises 91 scenes, with 76 for training and 15 for testing.
Each scene includes a pair of low-light and normal-light images along with their corresponding events.
\textbf{(2) SEE-600K} consists of 202 scenes, with each scene containing an average of four sets of videos under different lighting conditions, ranging from low light to bright light. 
During each training session, we randomly select one set of normal-light images as the reference and use the remaining sets as inputs. 
For example, for one scene with one low-light, two normal-light, and one high-light set, we generate six pairs of training data.

\subsection{Comparisons Experiments}
\textbf{Comparative Methods:} 
We categorize the competitive approaches into four groups. 
Firstly, DCE~\citeps{guo2020zero} is a classical approach that can adjust the image brightness curve to achieve normal lighting.
Secondly, there are strategies that only use images as input, including SNR~\citeps{xu2022snr}, UFormer~\citeps{wang2022uformer}, LLFlow~\citeps{wu2023learning}, and RetinexFormer~\citeps{cai2023retinexformer}. 
Thirdly, we consider methods that rely solely on events, \eg, E2VID+~\citeps{stoffregen2020reducing}.
Tertiary, we examine event-guided low-light enhancement frameworks.
This group includes single-frame input methods, \eg, eSL-Net~\citeps{wang2020event}, \citep{liu2023low}, \citep{wang2023event}and EvLight~\citeps{liang2024towards}, as well as multi-frame input strategies like EvLowLight~\citeps{liang2023coherent}.
Furthermore, we also compared the HDR reconstruction method HDRev~\citep{yang2023learning}.
We retrain all methods, following the open-source code when available; for approaches without open-source code, we replicate them based on their respective papers.

\noindent\textbf{Comparative on SDE Dataset:} 
The results from our comparative experiments, shown in Tab.~\ref{tab:sde_dataset} and Fig.~\ref{fig:6-SDE-VisualizationResults-Release}, reveal several key insights as following: 

\textit{\textbf{(1)} performance limitations of single-modal methods: }
Methods utilizing only a single modality exhibit limited performance, as shown in Tab.~\ref{tab:sde_dataset}. 
This trend highlights the need to integrate both modalities to achieve improved results, as demonstrated by methods like DCE \citeps{guo2020zero}, SRN \citeps{xu2022snr}, and UFormer \citeps{wang2022uformer} in Tab.\ref{tab:sde_dataset}.
The employment of a single modality does not supplement the dynamic range information, leading to relatively poorer performance, as shown in Fig.\ref{fig:6-SDE-VisualizationResults-Release}.

\textit{\textbf{(2)} effectiveness of event-guided methods: }
In contrast, event-guided image methods demonstrate significantly better performance.
Methods such as SEENet (Ours), EvLight, and others, as seen in Tab.~\ref{tab:sde_dataset}, demonstrate the best results. 
These approaches leverage the complementary strengths of both events and traditional images, leading to better outcomes in low-light conditions, as shown in Fig.~\ref{fig:6-SDE-VisualizationResults-Release} (g-j). 

\textit{\textbf{(3)} impact of indoor and outdoor conditions:}
Notably, performance in low-light indoor scenarios is inferior to that in outdoor settings, as shown in Fig.~\ref{fig:6-SDE-VisualizationResults-Release} (I) and (III).
This discrepancy may be attributed to the issues of flickering light sources commonly found indoors~\citeps{xu2023seeing}.
Our SEE-Net consistently achieves the best results across both scenarios, with a model size of just $1.9 M$—10\% parameter count of other SOTA methods—demonstrating its efficiency and compactness in low-light image enhancement.

\textit{\textbf{(4)} denoising is a core challenge: }
Fig.~\ref{fig:6-SDE-VisualizationResults-Release} present visualizations from the low-light outdoor scenes of the SDE dataset.
These low-light environments often come with significant noise, which poses a substantial challenge for current low-light enhancement methods. 
Our method demonstrates stable performance in addressing these noisy scenes, effectively enhancing the image quality while mitigating the noise, thereby highlighting the robustness of our approach in handling complex low-light conditions.

\noindent\textbf{Comparison on SEE-600K Dataset:}
The results presented in Tab.~\ref{tab:see_comp}, Fig.~\ref{fig:7-SEE-Dataset-Release}, and Fig.~\ref{fig:14-SEE-Dataset-More} showcase the performance of various methods on the SEE-600K dataset under different lighting conditions.

\textit{\textbf{(1)} Models trained on SDE:}
Models trained on the SDE dataset perform reasonably well when tested on the SEE-600K dataset, particularly under low-light conditions. Notably, the DCE~\citep{guo2020zero} method achieves the best performance in high-light scenarios, demonstrating its strong generalization capability, particularly for self-supervised learning approaches. However, in high-light and normal-light conditions, almost all methods fail, which highlights a significant limitation. This indicates that models trained on the SDE dataset learn a fixed multiplicative enhancement for illumination, and are not adaptable to a wider range of lighting conditions. This emphasizes the importance of redefining low-light enhancement as an illumination adjustment problem, as we have proposed.

\textit{\textbf{(2)} Models trained on SEE-600K:}
In contrast, models trained on the SEE-600K dataset show improved performance across both low-light and high-light conditions. Our proposed SEE-Net method outperforms other approaches, as shown in Tab.~\ref{tab:see_comp} and Fig.~\ref{fig:7-SEE-Dataset-Release}. This success can be attributed to our innovative use of prompt adjustments, which effectively resolve the ambiguities typically encountered in image enhancement processes. Further quantitative analysis confirms that SEE-Net consistently achieves superior PSNR, SSIM, and $L_1$ loss scores across lighting conditions, further validating the effectiveness of our approach.

\textit{\textbf{(3)} Advantages of prompt adjustments:}
Unlike previous methods that rely on one-way mapping, our approach with prompt adjustments demonstrates significant advantages, as shown in Fig.~\ref{fig:7-SEE-Dataset-Release} (h, p) and Fig. \ref{fig:14-SEE-Dataset-More} (h) of (I, II, III).
These adjustments allow us to generate images that surpass ground truth quality in both low-light and high-light conditions, particularly when the prompt is set to 0.5. In this setting, the output achieves optimal brightness and sharper textures.
Fig.~\ref{fig:7-SEE-Dataset-Release} and Fig.~\ref{fig:14-SEE-Dataset-More} further demonstrate the robustness and consistency of SEE-Net. 
Notably, SEE-Net is capable of producing more stable, high-quality images, with some outputs even surpassing the ground truth normal-light images.

However, there are still challenges, as illustrated in Fig.~\ref{fig:14-SEE-Dataset-More}.
In areas with complex textures or high-resolution requirements, all methods, including SEE-Net, struggle to achieve optimal results. Despite this, SEE-Net outperforms existing methods, particularly in terms of maintaining image quality and stability. These insights highlight the strengths of SEE-Net while also pointing out areas for potential improvement in future research.

\begin{table}[t]
\centering
\caption{Impact of Bayer Pattern Embedding. Removing the Bayer-pattern embedding results in a drop in PSNR and SSIM, showing its contribution to accuracy but not as the most critical factor.}
\resizebox{\linewidth}{!}{
\setlength{\tabcolsep}{0.1\linewidth}
\begin{tabular}{ccc}
\toprule
 \textbf{Bayer Pattern} & \textbf{PSNR} & \textbf{SSIM} \\
\midrule
 $f_{pe}$ & 23.57 & 0.7724 \\
 \textit{wo} & 22.94 & 0.7686 \\
\bottomrule
\end{tabular}
}
\label{tab:ablation_bayer}
\end{table}

\begin{table}[t]
\centering
\caption{Impact of Encoding Strategy. Replacing cross-attention with convolution-based approaches degrades performance, highlighting the importance of cross-attention.}
\resizebox{\linewidth}{!}{
\setlength{\tabcolsep}{0.1\linewidth}
\begin{tabular}{ccc}
\toprule
 \textbf{Encoding} & \textbf{PSNR} & \textbf{SSIM} \\
\midrule
 $f_{ca}$ & 23.57 & 0.7724 \\
 $add+conv$ & 22.40 & 0.7224 \\
 $cat+conv$ & 22.84 & 0.7298 \\
\bottomrule
\end{tabular}
}
\label{tab:ablation_encoding}
\end{table}

\begin{table}[t]
\centering
\caption{Impact of Loop Iterations. Reducing the number of loops from 20 to 10 lowers performance, showing that sufficient iterations are necessary for refinement.}
\resizebox{\linewidth}{!}{
\setlength{\tabcolsep}{0.1\linewidth}
\begin{tabular}{ccc}
\toprule
 \textbf{Loop Iterations} & \textbf{PSNR} & \textbf{SSIM} \\
\midrule
 20 & 23.57 & 0.7724 \\
 10 & 22.18 & 0.6812 \\
\bottomrule
\end{tabular}
}
\label{tab:ablation_loop}
\end{table}

\begin{table}[t]
\centering
\caption{Impact of Prompt Embedding. Replacing the learned embedding with a sine function achieves similar performance but does not surpass the learned embedding.}
\resizebox{\linewidth}{!}{
\setlength{\tabcolsep}{0.1\linewidth}
\begin{tabular}{ccc}
\toprule
 \textbf{Prompt Embedding} & \textbf{PSNR} & \textbf{SSIM} \\
\midrule
 $f_{pe}$ & 23.57 & 0.7724 \\
 $sin$ & 23.08 & 0.7692 \\
\bottomrule
\end{tabular}
}
\label{tab:ablation_prompt_embedding}
\end{table}

\begin{table}[t]
\centering
\caption{Impact of Prompt Merge. Disabling the prompt merge function slightly reduces performance, indicating its importance in optimizing results.}
\resizebox{\linewidth}{!}{
\setlength{\tabcolsep}{0.1\linewidth}
\begin{tabular}{ccc}
\toprule
 \textbf{Prompt Merge} & \textbf{PSNR} & \textbf{SSIM} \\
\midrule
 $add$ & 23.57 & 0.7724 \\
 $add$ (disabled) & 22.26 & 0.7713 \\
\bottomrule
\end{tabular}
}
\label{tab:ablation_prompt_merge}
\end{table}

\begin{table}[t]
\centering
\caption{Impact of Multi-Prompt Adjustment. Changing the prompt merge function from addition to multiplication yields the highest SSIM.}
\resizebox{\linewidth}{!}{
\setlength{\tabcolsep}{0.1\linewidth}
\begin{tabular}{ccc}
\toprule
 \textbf{Prompt Merge} & \textbf{PSNR} & \textbf{SSIM} \\
\midrule
 $add$ & 23.57 & 0.7724 \\
 $multiplication$ & 22.94 & 0.7893 \\
\bottomrule
\end{tabular}
}
\label{tab:ablation_multi_prompt}
\end{table}

\subsection{Ablation and Analytical Experiments}

\begin{figure*}[t!]
\centering
\includegraphics[width=0.98\linewidth]{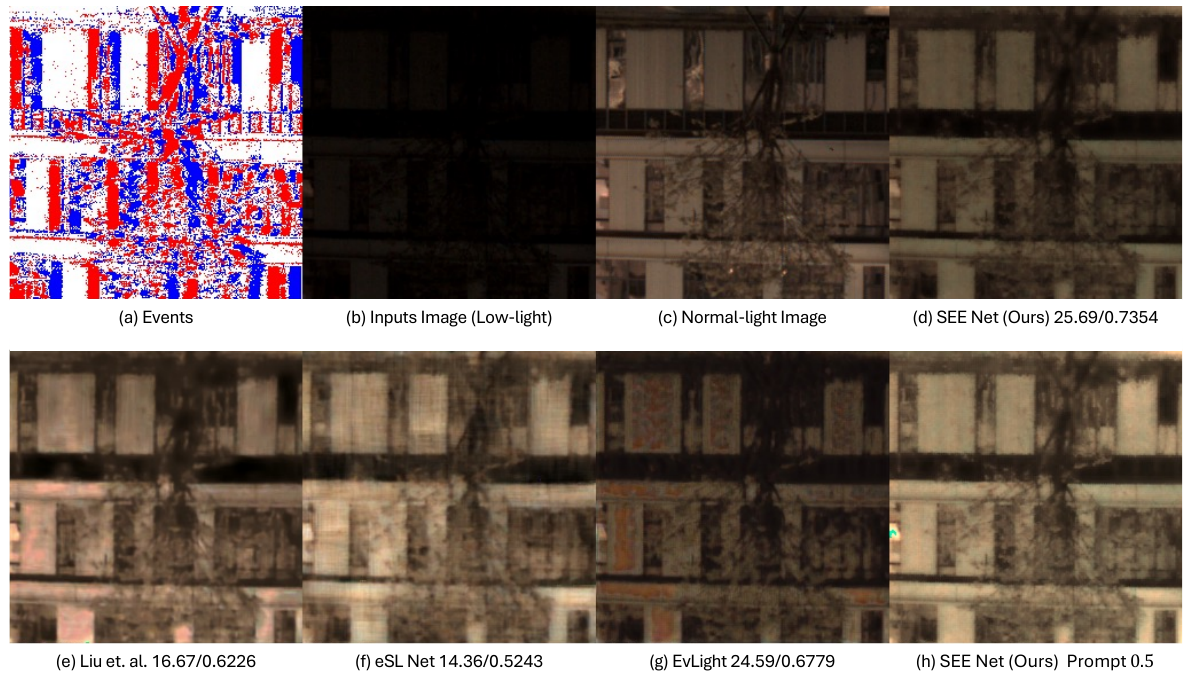}
\caption{Performance when there is relative motion between the foreground and the background.}
\label{fig:15-SeparationForegroundBackground}
\end{figure*}

In this ablation study, we analyze the impact of various components.

\textit{\textbf{(1)} bayer pattern embedding}:
Removing the Bayer-pattern embedding leads to a noticeable drop in both PSNR and SSIM, as shown in Tab.~\ref{tab:ablation_bayer}. The performance decline highlights that the Bayer pattern embedding contributes positively to the accuracy of the model, but it is not the most critical factor in performance. As seen in the table, the PSNR drops from 23.57 to 22.94 and the SSIM decreases from 0.7724 to 0.7686, indicating its moderate but important contribution.

\textit{\textbf{(2)} encoding strategy}: 
Replacing the cross-attention module $f_c$ with a convolution-based approach results in significant performance degradation, as seen in Tab.~\ref{tab:ablation_encoding}. The PSNR and SSIM decrease across both the `add` and `concat` convolution strategies compared to the use of cross-attention. Specifically, replacing cross-attention with the `add + conv` approach drops PSNR from 23.57 to 22.40 and SSIM from 0.7724 to 0.7224, while using `cat + conv` performs slightly better but still results in lower performance (22.84 PSNR and 0.7298 SSIM). These results emphasize the critical role of cross-attention in capturing complex features and enhancing image quality.

\textit{\textbf{(3)} loop iterations}: 
Reducing the number of loop iterations from 20 to 10 results in a noticeable performance drop, as reflected in Tab.~\ref{tab:ablation_loop}. The PSNR decreases from 23.57 to 22.18, and the SSIM drops from 0.7724 to 0.6812, indicating that sufficient iterations are necessary for effective refinement. These findings underline the importance of multiple iterations in optimizing image quality.

\textit{\textbf{(4)} prompt embedding}: 
Switching the prompt embedding from $f_{pe}$ to a sine function results in similar performance, but it does not surpass the learned embedding. As shown in Tab.~\ref{tab:ablation_prompt_embedding}, while both embeddings yield similar PSNR and SSIM, the learned embedding outperforms the sine function method in terms of image quality, achieving a PSNR of 23.57 and SSIM of 0.7724 compared to 23.08 PSNR and 0.7692 SSIM for the sine function. This shows that while sine functions provide comparable results, the learned embedding offers a slight performance advantage.

\textit{\textbf{(5)} prompt merge}: 
Disabling the prompt merge results in a slight performance decrease, as shown in Tab\ref{tab:ablation_prompt_merge}. The PSNR drops from 23.57 to 22.26, and the SSIM decreases from 0.7724 to 0.7713, demonstrating that prompt merging is important for optimizing results. This suggests that prompt merging plays a role in effectively integrating the image and event features.

\textit{\textbf{(6)} multi-prompt adjustment}: 
Fig.~\ref{fig:8-Prompt0.3to0.7} demonstrates the effect of multi-prompt adjustments. When using gamma correction to brighten the low-light input, significant noise is introduced, as shown in Fig.~\ref{fig:8-Prompt0.3to0.7} (a). However, by adjusting the prompt, we can effectively control the brightness while reducing noise, as seen in Fig.~\ref{fig:8-Prompt0.3to0.7} (b). These results highlight the flexibility and robustness of our method in post-processing, where multiple prompt adjustments provide fine-grained control over image enhancement.

\textit{\textbf{(7)} Generalization of Relative Motion:}
Although our dataset is designed primarily to capture static scenes, it inherently includes relative motion between objects.
For instance, in a three-dimensional space, the nonlinear movement of the robotic arm causes objects at varying distances from the camera to move at different speeds, thus introducing motion in the image, as demonstrated in Fig. \ref{fig:15-SeparationForegroundBackground}.
In real-world scenarios, such relative motion is common, especially in dynamic environments. 
This relative motion enables our method to model local movements within the scene. 
This characteristic is crucial because it ensures that our method can generalize effectively to scenarios where objects in the scene move at varying speeds relative to the camera.

\section{Conclusion}
In this paper, we proposed a new research problem: how to use events to adjust the brightness of images across a wide range of lighting conditions, from low light to high light. 
To address this challenge, we made the following contributions. 
\begin{itemize}
    \item \textbf{(1)}, we developed a physical model and formally defined the problem of brightness adjustment using events, providing a solid theoretical foundation. 
    \item \textbf{(2)}, we introduced a new spatiotemporal registration algorithm based on a robotic arm and collected a large-scale dataset, \textbf{SEE-600K}, to overcome alignment issues and support our research. 
    \item \textbf{(3)}, we presented \textbf{SEE-Net}, a novel and compact framework capable of accepting input images with a wide range of illumination and producing output images with adjustable brightness. 
    \item \textbf{(4)}, we conducted extensive experiments to demonstrate the effectiveness of our method.
\end{itemize}

\noindent \textbf{Limitation and Future Works}: Since this is the first attempt to address this novel problem, brightness adjustment with events, our method has not yet reached a level suitable for direct commercial application, and further improvements are needed in future research.

\section*{Supplementary Materials Overview}

In the supplementary materials, we provide a more detailed introduction to the imaging process of the images in Appendix A, and also analyze how events assist in adjusting the imaging process. 
Since this section involves many fundamental concepts and models related to the imaging process, we expand on them in the Appendix.

Additionally, the detailed explanation and derivation of the IMU algorithm are also presented in the supplementary materials, specifically in Appendix B, where Kalman filtering is used to process the IMU data and subsequently design the registration algorithm.
We hope that this registration method can also be considered for use with other datasets.

In the supplementary video, we showcase additional examples from the dataset, as well as the results of different methods applied to the videos. Additionally, we also present a visualization of the registration process, making it easier for readers to understand our IMU registration algorithm.

\section*{Data Availability Statement}

The datasets supporting the results of this article are publicly available. 
The SDE dataset used in this study can be accessed at \url{https://github.com/EthanLiang99/EvLight}.
The SEE-600K dataset is available at \url{https://github.com/yunfanLu/SEE}.
We commit to making all data and code related to this work openly accessible.

\begin{acknowledgements}
This work was supported in part by the National Natural Science Foundation of China (Grant No.92370204),in part by the National Key R\&D Program of China (Grant No.2023YFF0725001), in part by Guangzhou-HKUST(GZ) Joint Funding Program (Grant No.2023A03J0008), in part by the Education Bureau of Guangzhou Municipality.
\end{acknowledgements}

{\small
\bibliographystyle{spbasic}
\bibliography{iclr2025_conference}
}


\end{document}